\documentclass[sigconf,nonacm]{acmart}
\AtBeginDocument{%
  }

\setcopyright{acmlicensed}
\copyrightyear{2025}
\acmYear{2025}
\acmConference[Web of Data Quality Workshop at the ACM Web Conference]{Web of Data Quality Workshop at the ACM Web Conference}{Apr 28--May 02, 2025}{Sydney, Australia}
\acmBooktitle{ACM Web Conference '25: Companion Proceedings,
  Apr 28--May 02, 2025, Sydney, Australia}

\usepackage{listings}
\usepackage{soul}
\usepackage{cleveref}
\usepackage[inline]{enumitem}
\usepackage{subcaption}
\usepackage[ruled,vlined]{algorithm2e} 

\usepackage{titlesec}

\begin{document}

\title{Explainability-Driven Quality Assessment for Rule-Based Systems}

\author{Oshani Seneviratne}
\email{senevo@rpi.edu}
\affiliation{%
  \institution{Rensselaer Polytechnic Institute}
  \city{Troy}
  \state{New York}
  \country{USA}
}

\author{Brendan Capuzzo}
\email{capuzb@rpi.edu}
\affiliation{%
  \institution{Rensselaer Polytechnic Institute}
  \city{Troy}
  \state{New York}
  \country{USA}
}

\author{William Van Woensel}
\email{VanWoensel@telfer.uOttawa.ca}
\affiliation{%
  \institution{University of Ottawa}
  \city{Ottawa}
  \country{Canada}
}

\renewcommand{\shortauthors}{Seneviratne et al.}

\begin{abstract}
This paper introduces an explanation framework designed to enhance the quality of rules in knowledge-based reasoning systems based on dataset-driven insights. The traditional method for rule induction from data typically requires labor-intensive labeling and data-driven learning. This framework provides an alternative and instead allows for the data-driven refinement of existing rules: it generates explanations of rule inferences and leverages human interpretation to refine rules. It leverages four complementary explanation types—trace-based, contextual, contrastive, and counterfactual—providing diverse perspectives for debugging, validating, and ultimately refining rules. By embedding explainability into the reasoning architecture, the framework enables knowledge engineers to address inconsistencies, optimize thresholds, and ensure fairness, transparency, and interpretability in decision-making processes. Its practicality is demonstrated through a use case in finance. 
\end{abstract}


\maketitle

\section{Introduction}
\label{sec:introduction}

The reliability, utility, and trustworthiness of knowledge-based systems, and, ultimately, the \emph{Web of Data}, lies in the quality of the underlying logic-based rules. Rules that are inaccurate or incomplete, or, in general, do not capture their intent, compromise the effectiveness of reasoning systems that rely on them. 
To improve rule quality in a data-driven way, machine learning techniques such as rule induction or decision trees can be used to extract rules from data. However, these methods typically require labeled datasets, which are labor-intensive to prepare; also, the output is a new set of rules that needs to be compared to the current ones.
This paper instead introduces an explanation-driven framework for refining existing rules, 
which enhances reliability, transparency, and trust in reasoning systems without the effort of full-scale data labeling.

To that end, the framework generates explanations of rule inferences, which can be interpreted by knowledge engineers to refine the existing rules. 
Our framework leverages four complementary explanation types (trace-based, contextual, contrastive, and counterfactual) to provide diverse perspectives for debugging, validating, and ultimately refining rules.
These explanation types empower users to understand the derivation of conclusions, explore their immediate and upstream causes, compare alternative scenarios, and, in doing so, generate actionable insights for improvement. 
We integrated these explanation mechanisms into Punya~\cite{patton2019app, patton2021punya}, a Semantic Web fork of the MIT App Inventor platform, which is a low-code platform for app development. This choice was made to support both technical and non-technical knowledge engineers in debugging and refining rules in a data-driven way. Punya, in particular, was chosen due to its built-in support for rule-based reasoning.
The proposed explanation framework primarily benefits knowledge engineers; by enhancing rule transparency through explanations, the framework can also benefit end-users who interact with reasoning-driven applications.
%
To the best of our knowledge, our work is among the first to explicitly investigate the use of explanations for refining rules in knowledge-based systems. 



The remainder of this paper is structured as follows. \Cref{sec:related-work} discusses related work on rule quality and debugging, highlighting how our approach complements and extends prior efforts. In \Cref{sec:reasoning_architecture}, we describe the reasoning architecture underpinning the explanation framework. \Cref{sec:use_cases} presents a compelling use case that benefits from our solution. \Cref{sec:explanations_implementation} demonstrates the explanation types, illustrating their role in enhancing rule quality. \Cref{sec:integration_testing} outlines the integration of the framework into the MIT App Inventor (Punya) platform, emphasizing its features, lightweight implementation, and practical benefits. Finally, \Cref{sec:discussion} and \Cref{sec:conclusion} discuss the implications of this work and conclude with future directions for improving data quality and explainability in knowledge-based systems.

\section{Related Work}
\label{sec:related-work}

The quality and debugging of rules in knowledge-based systems have been studied in several contexts, with efforts ranging from foundational correctness principles to advanced explanation frameworks. This section reviews key contributions and situates our work within this landscape.

\subsection{Rule Quality Principles}

\citet{landauer1990correctness} propose a set of acceptability principles—Consistency, Completeness, Irredundancy, Connectivity, and Distribution—to guide the construction and validation of rule bases. These principles provide foundational criteria for evaluating rule systems, emphasizing logical soundness, efficiency, and simplicity. 
In this vein, we focus on communicating the behavior of rules to knowledge engineers or end-users by employing user-centric explanations— trace-based, contextual, contrastive, and counterfactual—to concretely help with identifying inconsistencies, incompleteness, accuracy, and fairness of decision-making.


\subsection{Debugging and Validation Techniques}

In the domain of declarative programming, \citet{gebser2008meta} introduces a meta-programming technique for debugging answer-set programming (ASP). Their approach allows querying why a single interpretation, or class of interpretations, is not an answer set for a program.
This method identifies semantical errors within the context of ASP and can be used for debugging answer-set programs.
However, their work is inherently tied to the semantics of ASP and does not extend to broader rule-based reasoning systems. 
In contrast, our framework generalizes debugging and validation techniques through the use of a number of explanation types.
By focusing on explanation-driven rule refinement rather than program-level debugging, our work moves from identifying errors within a specific programming paradigm to enhancing the quality of rules that underpin knowledge-based reasoning in general. 

\subsection{Explanation Frameworks}

Explanation frameworks have emerged as critical tools for enhancing the interpretability and usability of AI systems. Explanation Ontology (EO)~\cite{chari2020explanation, chari2023explanation} provides a general-purpose semantic framework for representing explanations, supporting 15 distinct explanation types with clear definitions and logical formalizations using Web Ontology Language (OWL). EO enables system designers to connect explanations to underlying data and reasoning processes, ensuring that AI systems address user-centered needs effectively. Practical applications of EO include contextual explanations in healthcare, such as helping clinical practitioners understand AI-driven risk predictions for Type-2 Diabetes and Chronic Kidney Disease (CKD)~\cite{chari2023informing}. While EO provides a robust theoretical foundation for explainability, our framework focuses on practical implementation within the MIT App Inventor Punya platform, demonstrating its utility through real-world applications and diverse explanation types.

Similarly, XAIN (eXplanations for AI in Notation3)~\cite{vanwoensel2023explanations} supports trace-based, contrastive, and counterfactual explanations, specifically focusing on healthcare applications. For example, XAIN explains recommendations for Chronic Obstructive Pulmonary Disease (COPD) patients in order to effect understanding, persuasion, and behavior change. 
While XAIN showcases the power of symbolic reasoning for generating explanations, it does not address rule quality as a central theme. Our work builds upon these advancements by implementing explanations that explicitly target rule validation and debugging, 
and integrating these capabilities into a lightweight reasoning architecture for mobile and resource-constrained environments, as described in the following section.


\section{Reasoning Architecture}
\label{sec:reasoning_architecture}

\subsection{MIT App Inventor and Punya}

MIT App Inventor is a web-based, drag-and-drop platform designed to make app development accessible to users with minimal programming experience~\cite{patton2019app, patton2021punya}. Punya builds upon this foundation by introducing support for \emph{Linked Data} and lightweight reasoning, enabling developers to create sophisticated applications that leverage semantic technologies~\cite{li2019semantic}.
By providing drag-and-drop tools and predefined components for semantic web data creation and consumption, Punya reduces the technical burden on developers, enabling even non-experts to create mobile apps with semantic features and reasoning capabilities~\cite{seneviratne2014developing}.
Furthermore, Punya’s advanced capabilities make it a valuable tool for researchers and professionals in domains such as life sciences, disaster response, and healthcare. For example, Punya has been used to build disaster relief apps that crowdsource reports in real time~\cite{shih2013democratizing} and healthcare apps that enable data-driven self-management for chronic conditions~\cite{patton2022development}.

Punya's main features include:

\begin{itemize}
    \item \textbf{Linked Data Support:} Punya allows mobile apps to interlink structured data with semantic data to form RDF triples, thereby enabling seamless integration with external Linked Data sources and enhancing data interoperability. 

    \item \textbf{Reasoner Component:} The platform includes a lightweight reasoning engine capable of operating on individual mobile devices. Built on Apache Jena’s RDF reasoning framework, Punya’s reasoner enables apps to process knowledge graphs and apply inference rules to derive new facts.
\end{itemize}

\begin{figure*}[!htbp] 
    \centering
    \includegraphics[width=\textwidth]{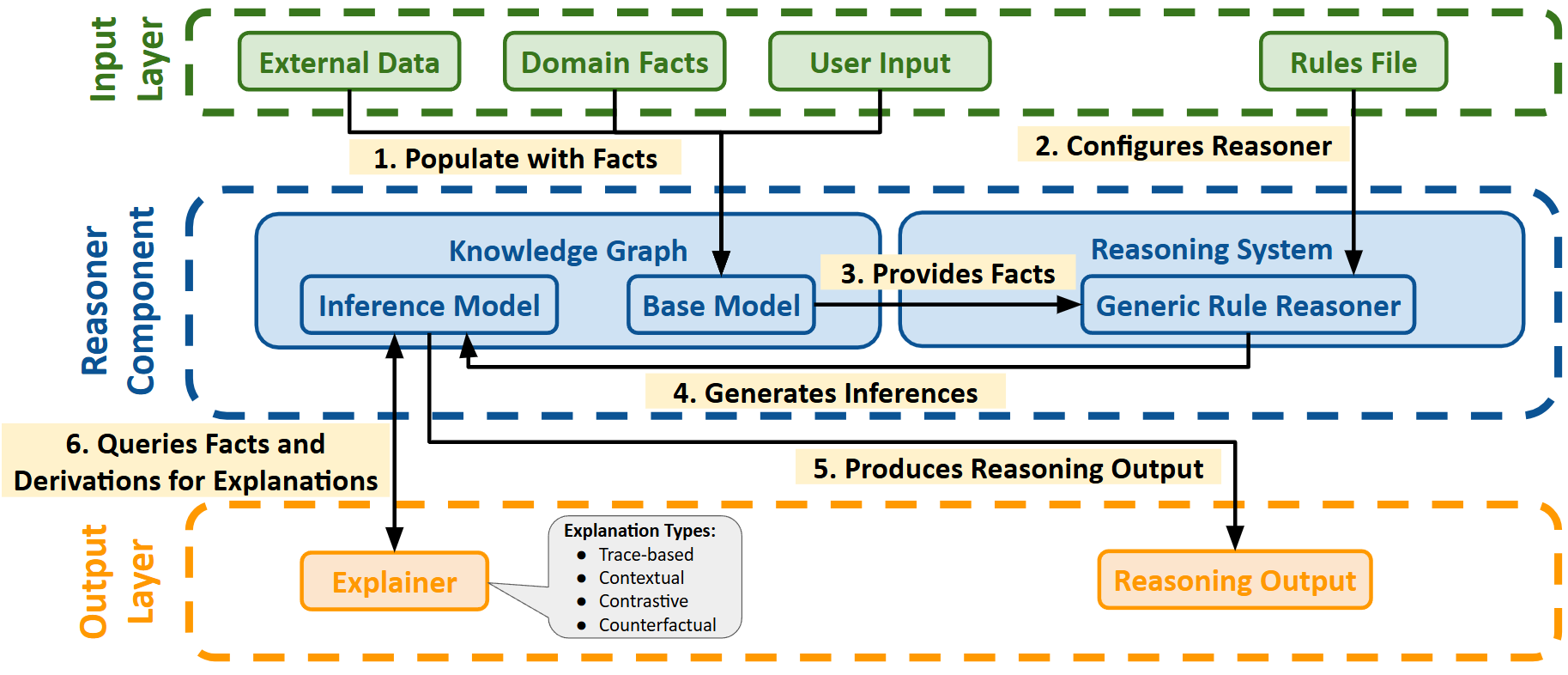}
    \caption{Reasoning Architecture with Explainer Component}
    \label{fig:reasoner_architecture}
\end{figure*}

In this work, the modifications to the reasoning architecture of the Punya platform with the \emph{Explainer Component} are depicted in Figure~\ref{fig:reasoner_architecture}. It is designed to facilitate the derivation of new knowledge from input data and rules while offering explanations to end-users. It consists of three key layers: the Input Layer, the Reasoner Component, and the Output Layer, as explained below.

\subsection{Input Layer}

The Input Layer is responsible for initializing the reasoning process by supplying data inputs to the system, such as:

\begin{itemize}
    \item \textbf{Domain Facts:} Context-specific information required by the application.
    \item \textbf{Rules File:} A collection of inference rules written in Jena's rule syntax. 
    \item \textbf{User Input:} Personalized data provided by the user.
    \item \textbf{External Data:} Structured data sources containing auxiliary data needed for the application, such as a SPARQL endpoint or an RDF data file.
\end{itemize}

\subsection{Reasoner Component}

The Reasoner Component is the core of the architecture and is composed of two primary subcomponents:

\begin{itemize}
    \item \textbf{Knowledge Graph:} The repository of all system knowledge, consisting of two models:
        \begin{itemize}
            \item \textbf{Base Model:} Stores initial facts and user-provided information.
            \item \textbf{Inference Model:} Extends the base model with newly reasoned facts and their derivation chains, representing the enriched knowledge state.
        \end{itemize}
    \item \textbf{Reasoning System:} Powered by an Apache Jena-based rule reasoner. This system applies inference rules to the knowledge graph to derive new facts by iteratively matching input data against rule conditions, generating and storing inferred triples along with their logical derivations in the inference model.
\end{itemize}

\subsection{Output Layer}

The Output Layer presents the results of the reasoning process in two forms:

\begin{itemize}
    \item \textbf{Reasoning Output:} Inferred facts (represented as RDF triples) derived from combining base facts and inference rules.
    \item \textbf{Explainer:} This module extracts derivation chains from the inference model to generate human-readable explanations. The explainer offers various types of explanations, such as trace-based, contrastive, and contextual, to provide end-users with an understanding of how and why specific conclusions were reached.
\end{itemize}

This dual-output design ensures that both technical and lay users can access and comprehend the reasoning process and its results.

\subsection{Explanation Component}
\label{sec:explanation_component}

\begin{figure}[h!]
    \centering
    \includegraphics[width=\columnwidth]{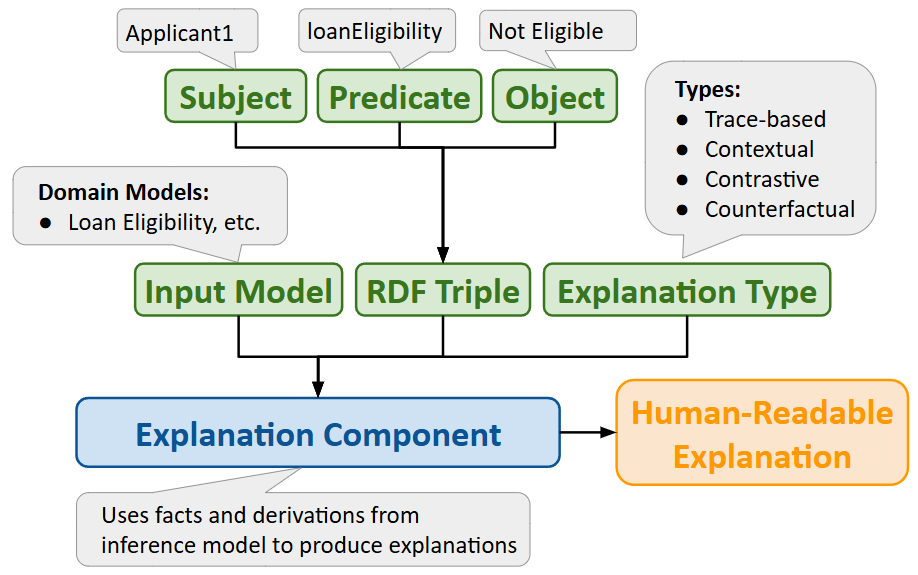}
    \caption{The Explanation Component and its Inputs}
    \label{fig:explanation_component}
\end{figure}

The explanation component is a central element in the reasoning architecture, designed to transform inferred knowledge into human-readable explanations tailored to user needs. As illustrated in \Cref{fig:explanation_component}, the component processes three key inputs to generate explanations: the input model, the statement (i.e., the RDF triple we want an explanation for), and the explanation type. 

\paragraph{Inputs to the Explanation Component}

\begin{itemize}
    \item \textbf{Input Model:} The input model represents the domain-specific knowledge base used in the explanation process. It provides the contextual framework within which the inferred facts are analyzed and explained.
    
    \item \textbf{Statement:} The statement, in the form of an RDF triple, serves as the focal point of the explanation. It represents the inferred fact generated by the reasoning component. Users select the statement they wish to investigate, enabling the component to trace its derivation and provide an explanation of how the conclusion was reached.

    \item \textbf{Explanation Type:} The explanation type specifies the format or approach used to present the information to the user. This module currently supports trace-based, contrastive, and contextual explanations, each tailored to different user needs and scenarios. The choice of explanation type determines the level of detail and focus in the generated output.
\end{itemize}

\paragraph{Functionality of the Explanation Component}

The explanation component operates by querying the inference model for relevant facts and derivations associated with the selected RDF triple. Using this data, it generates the desired explanation that aligns with the chosen explanation type, leveraging different strategies as illustrated in \Cref{sec:explanations_implementation}. 

\section{Use Case}
\label{sec:use_cases}


To evaluate the utility of the proposed explanation component, we applied it to a financial decision-making use case.

Here, a financial system assesses loan eligibility based on credit scores and debt-to-income (DTI) ratios.

\begin{itemize}
    \item \textbf{Base Facts:} 
        \begin{itemize}
            \item Alex: Credit Score 680, Monthly Debt \$2000, Monthly Income \$5000.
            \item Beth: Credit Score 605, Monthly Debt \$1500, Monthly Income \$5000.
            \item Charlie: Credit Score 700, Monthly Debt \$1000, Monthly Income \$5000.
        \end{itemize}
    \item \textbf{Rules:}
        \begin{itemize}
            \item Calculate DTI ratio: \texttt{[DTIRule]}
            \item Determine eligibility: \texttt{[EligibilityRule]}, \\\texttt{[NotEligibleDTIRule]}, and \texttt{[NotEligibleCreditRule]}.
        \end{itemize}
    \item \textbf{Inferred Facts:}
        \begin{itemize}
            \item Alex: DTI 0.40 (Not Eligible - High DTI).
            \item Beth: DTI 0.30 (Not Eligible - Low Credit Score).
            \item Charlie: DTI 0.20 (Eligible - Good DTI and Credit Score).
        \end{itemize}
\end{itemize}

The Explanation Component explains why specific applicants are eligible or ineligible, detailing the reasoning behind inferred DTI ratios and eligibility outcomes. 

\section{Explanation Types Supported}
\label{sec:explanations_implementation}

\subsection{Trace-Based Explanations}

Trace-based explanations provide a comprehensive account of the reasoning process by detailing the complete derivation chain for a given conclusion. This explanation type is valuable for understanding how facts trigger rules, how rules interact among themselves, and checking the correctness of inferences in knowledge-based systems. By presenting a step-by-step breakdown of the applied rules and supporting facts, trace-based explanations serve multiple purposes:
\begin{enumerate*}[label=(\roman*)]
    \item Detecting and diagnosing unexpected rule interactions that led to unexpected conclusions.
    \item Ensuring the consistency and accuracy of inferred facts with respect to input data and predefined rules.
    \item Verifying the integrity of complex rule chains, i.e., in scenarios involving cascading inferences.
\end{enumerate*}

The algorithm for generating trace-based explanations, presented in Algorithm \ref{alg:trace_based_explanation}, outlines the systematic process of identifying derivations, evaluating rule applications, and formatting the explanation for user interpretation. 


\begin{algorithm}[h]
\SetAlgoLined
\caption{Trace-Based Explanation Generation}
\label{alg:trace_based_explanation}

\SetKwFunction{CreateExplainer}{CreateExplainer}
\SetKwFunction{GenerateTrace}{GenerateTrace}
\SetKwFunction{ListStatements}{ListStatements}
\SetKwFunction{CheckAssertion}{CheckAssertion}
\SetKwFunction{FindMatchingRules}{FindMatchingRules}
\SetKwFunction{IdentifySupportingFactsRules}{IdentifySupportingFactsRules}
\SetKwFunction{FormatExplanation}{FormatExplanation}

\SetKwBlock{Main}{Main Procedure}{end}
\SetKwProg{Fn}{Function}{:}{end}

\Fn{\GenerateTrace{$M$}}{
    \tcp{List RDF triples from inference model $M$}
    $S \gets$ \ListStatements{$M$} \\
    $E \gets \emptyset$ \tcp*{Initialize explanation}

    \ForEach{$s \in S$}{
        \eIf{\CheckAssertion{$s$}}{
            \tcp{Statement $s$ is directly asserted}
            Add $s$ to $E$
        }{
            \tcp{Find rules that inferred $s$}
            $R \gets$ \FindMatchingRules{$s$} \\
            \tcp{Recursively identify supporting facts \& rules}
            $F \gets$ \IdentifySupportingFactsRules{$s, R$}
            Add $s, R, F$ to $E$
        }
    }
    \textbf{return} \FormatExplanation{$E$}
}
\end{algorithm}

In the loan eligibility model, a trace-based explanation can show that an applicant was denied a loan because their DTI ratio exceeded the acceptable threshold (\Cref{lst:trace_based_explanation}). The explanation traces the reasoning process from the conclusion to the supporting facts and rules applied, highlighting the interactions between the DTI ratio and loan eligibility rules.

\begin{lstlisting}[
    float=htbp,
    label={lst:trace_based_explanation},
    caption={Trace-based explanation illustrating the derivation of loan ineligibility for \texttt{applicant1}. },
    basicstyle=\scriptsize\ttfamily,
    breaklines=true,
    columns=fullflexible,
    frame=single,
    backgroundcolor=\color{gray!10},
    xleftmargin=1em,
    xrightmargin=0.5em,
    language={}
]
Conclusion: applicant1 has Loan Eligibility: Not Eligible
  Match: applicant1 has Type: Person
  Match: applicant1 has DTI Ratio: 0.4
  
  Conclusion: applicant1 has DTI Ratio: 0.4
    Match: applicant1 has Type: Person
    Match: applicant1 has Monthly Debt: 2000.0
    Match: applicant1 has Monthly Income: 5000.0
    Rule: [DTIRule: 
      (?applicant type Person)
      (?applicant monthlyDebt ?debt)
      (?applicant monthlyIncome ?income)
      quotient(?debt ?income ?dti)
      -> (?applicant dtiRatio ?dti)]

  Rule: [NotEligibleDTIRule:
    (?applicant type Person)
    (?applicant dtiRatio ?dti)
    greaterThan(?dti '0.349999')
    -> (?applicant loanEligibility 'Not Eligible')]
\end{lstlisting}

\subsection{Contextual Explanations}

Contextual explanations offer a targeted view of the reasoning process by focusing on the immediate rule and supporting facts responsible for a specific inference. Unlike trace-based explanations, which detail the entire reasoning chain, contextual explanations emphasize only the most relevant subset of information, making them particularly effective for:
\begin{enumerate*}[label=(\roman*)]
    \item Streamlining the debugging process by isolating localized issues without overwhelming users with extraneous details.
    \item Facilitating the validation of individual rules by clearly identifying their direct contribution to a given conclusion.
    \item Enhancing communication with non-technical stakeholders by presenting concise, focused insights into rule behavior.
\end{enumerate*}

Algorithm~\ref{alg:contextual_explanation_generation} outlines the step-by-step process of generating contextual explanations.

\begin{algorithm}[h]
\SetAlgoLined
\caption{Contextual Explanation Generation}
\label{alg:contextual_explanation_generation}

\SetKwFunction{CreateExplainer}{CreateExplainer}
\SetKwFunction{GenerateContextualExplanation}{GenerateContextualExplanation}
\SetKwFunction{AccessDerivations}{AccessDerivations}
\SetKwFunction{IdentifyTriggeredRule}{IdentifyTriggeredRule}
\SetKwFunction{ExtractMatchingFacts}{ExtractMatchingFacts}
\SetKwFunction{FormatExplanation}{FormatExplanation}

\SetKwBlock{Main}{Main Procedure}{end}
\SetKwProg{Fn}{Function}{:}{end}

\Fn{\GenerateContextualExplanation{$M$}}{
    \tcp{Access derivations from inference model $M$}
    $D \gets$ \AccessDerivations{$M$} \\
    $E \gets \emptyset$ \tcp*{Initialize explanation}

    \ForEach{$d \in D$}{
        \tcp{Identify rule that triggered conclusion}
        $R \gets$ \IdentifyTriggeredRule{$d$} \\
\tcp{Extract facts matching rule conditions}
        $F \gets$ \ExtractMatchingFacts{$d, R$} \\
        \tcp{Include rule, facts in explanation}
        Add $d, R, F$ to $E$
    }
    \tcp{Format in shallow, simplified form}
    \textbf{return} \FormatExplanation{$E$} 
}
\end{algorithm}

A contextual explanation can clarify why an applicant was deemed ineligible by showing that their DTI ratio (0.4) exceeded the threshold specified in the \texttt{NotEligibleDTIRule} (\Cref{lst:contextual_explanation}). The shallow explanation highlights the specific rules and matched facts, while the simplified version translates these insights into user-friendly language.

\begin{lstlisting}[
    float=htbp,
    caption={Contextual explanation for loan ineligibility of \texttt{applicant1}, showcasing both a shallow technical explanation and a simplified natural language explanation.},
    label={lst:contextual_explanation},
    basicstyle=\scriptsize\ttfamily,
    breaklines=true,
    columns=fullflexible,
    frame=single,
    backgroundcolor=\color{gray!10},
    xleftmargin=1em,
    xrightmargin=0.5em,
    language={}
]
Shallow Explanation:
Conclusion: applicant1 has Loan Eligibility: Not Eligible
Based on rule: [ NotEligibleDTIRule: (?applicant type Person) (?applicant dtiRatio ?dti) greaterThan(?dti '0.349999' -> (?applicant loanEligibility 'Not Eligible') ]
Using the following facts:
- applicant1 has Type: Person
- applicant1 has DTI Ratio: 0.4

Simple Explanation:
applicant1 has Loan Eligibility: Not Eligible because applicant1 has Type: Person and applicant1 has DTI Ratio: 0.4.
\end{lstlisting}

\subsection{Contrastive Explanations}

Contrastive explanations outline the differences between two cases with distinct outcomes, providing a comparative lens to understand how variations in data influence reasoning processes. 
By juxtaposing cases with contrasting results, this explanation type allows: 
\begin{enumerate*}[label=(\roman*)]
    \item Diagnosing subtle rule interactions or inconsistencies that lead to divergent outcomes in similar scenarios.
    \item Refining and calibrating rule thresholds to ensure fairness and consistency in different cases.
    \item Identifying data features that have a disproportionate influence on decision-making in different cases, thereby improving transparency and fairness.
\end{enumerate*}

By pinpointing the factors that contribute to differing outcomes, contrastive explanations empower knowledge engineers to refine rules and validate their robustness. The systematic procedure for generating contrastive explanations is detailed in Algorithm~\ref{alg:contrastive_explanation_generation}.

\begin{algorithm}[h]
\SetAlgoLined
\caption{Contrastive Explanation Generation}
\label{alg:contrastive_explanation_generation}

\SetKwFunction{CreateExplainer}{CreateExplainer}
\SetKwFunction{GenerateContrastiveExplanation}{GenerateContrastiveExplanation}
\SetKwFunction{GenerateInferenceModel}{GenerateInferenceModel}
\SetKwFunction{ProcessStatements}{ProcessStatements}
\SetKwFunction{IdentifyOutcomes}{IdentifyOutcomes}
\SetKwFunction{CompareModels}{CompareModels}
\SetKwFunction{FormatExplanation}{FormatExplanation}

\SetKwBlock{Main}{Main Procedure}{end}
\SetKwProg{Fn}{Function}{:}{end}

\Fn{\GenerateContrastiveExplanation{$B$, $C$}}{
    \tcp{Generate inference model for base model}
    $M_B \gets$ \GenerateInferenceModel{$B$} \\
    \tcp{Generate inference model for contrastive model}
    $M_C \gets$ \GenerateInferenceModel{$C$} \\
    \tcp{Identify outcomes, supporting facts for $B$}
    $O_B, F_B \gets$ \ProcessStatements{$M_B$} \\
    \tcp{Identify outcomes, supporting facts for $C$}
    $O_C, F_C \gets$ \ProcessStatements{$M_C$} \\
    \tcp{Identify similarities $S$, differences $D$ between models}
    $S, D \gets$ \CompareModels{$O_B, O_C, F_B, F_C$} 
    
    \tcp{Format explanation with similarities, differences}
    \textbf{return} \FormatExplanation{$S$, $D$}
}
\end{algorithm}

In the loan eligibility model, a contrastive explanation compares two applicants: one eligible for a loan and one ineligible. The explanation highlights differences in their DTI ratios, monthly debts, and credit scores and illustrates how these variations may interact with the rules to produce different outcomes (\Cref{lst:contrastive_explanation}).

\begin{lstlisting}[
    float=htbp,
    caption={Contrastive explanation comparing two applicants with differing loan eligibility outcomes.},
    label={lst:contrastive_explanation},
    basicstyle=\scriptsize\ttfamily,
    breaklines=true,
    columns=fullflexible,
    frame=single,
    backgroundcolor=\color{gray!10},
    xleftmargin=1em,
    xrightmargin=0.5em,
    language={}
]
Similarities:
    - applicant1 has Monthly Income: 5000.0

Differences:
    - For Monthly Debt: this model has 2000.00 while the alternate model has 1000.00
    - For Loan Eligibility: this model has Not Eligible while the alternate model has Eligible
    - For Credit Score: this model has 680 while the alternate model has 700
    - For DTI Ratio: this model has 0.40 while the alternate model has 0.20
\end{lstlisting}

\subsection{Counterfactual Explanations}

Counterfactual explanations leverage ``what-if" scenarios, offering insights into how changes to input data could result in different reasoning outcomes. 
This type of explanation offers actionable modifications to change the reasoning outcome of a given case. 
In doing so, this explanation type serves as a tool for:
\begin{enumerate*}[label=(\roman*)]
    \item Providing actionable guidance for improving input data to achieve desired outcomes.
    \item Adjusting rule conditions to achieve desired outcomes.
\end{enumerate*}

Algorithm~\ref{alg:counterfactual_explanation_generation} details the process for creating counterfactual explanations, leveraging a nearest-neighbor approach inspired by the NICE algorithm \cite{brughmans2024nice}.


\begin{algorithm}[h]
\SetAlgoLined
\caption{Counterfactual Explanation Generation}
\label{alg:counterfactual_explanation_generation}

\SetKwFunction{InitializeExplainer}{InitializeExplainer}
\SetKwFunction{GenerateCounterfactualExplanation}{GenerateCounterfactualExplanation}
\SetKwFunction{QueryHistoricalCases}{QueryHistoricalCases}
\SetKwFunction{CalculateFeatureDistance}{CalculateFeatureDistance}
\SetKwFunction{GenerateMinimalChanges}{GenerateMinimalChanges}
\SetKwFunction{ValidateProposedChanges}{ValidateProposedChanges}
\SetKwFunction{FormatCounterfactualExplanation}{FormatCounterfactualExplanation}
\SetKwFunction{IdentifyRelevantDifferences}{IdentifyRelevantDifferences}

\SetKwBlock{Main}{Main Procedure}{end}
\SetKwProg{Fn}{Function}{:}{end}

\Fn{\GenerateCounterfactualExplanation{$C$, $M$}}{
    \tcp{Retrieve historical ``approved'' cases}
    $H \gets$ \QueryHistoricalCases{$M$} \\
    \tcp{Initialize set of nearest neighbors}
    $N \gets \emptyset$

    \ForEach{$h \in H$}{
        \tcp{Calculate distance between current (non-approved) and historical cases}
        $d \gets$ \CalculateFeatureDistance{$C$, $h$} \\
        Add $(h, d)$ to $N$
    }
    \tcp{Identify nearest unlike neighbor}
    Sort $N$ by ascending $d$ \\
    $C^\prime \gets$ \IdentifyRelevantDifferences{$C$, $N[0]$} \\

    \If{\ValidateProposedChanges{$C^\prime$, $M$}}{
        \tcp{Format counterfactual explanation}
        \textbf{return} \FormatCounterfactualExplanation{$C^\prime$} 
    }
    \tcp{No valid counterfactual explanation found}
    \textbf{return} Failure
}

\Fn{\IdentifyRelevantDifferences{$C$, $h$}}{
    \tcp{Select features where differences significantly impact the outcome}
    $D \gets \emptyset$ \\
    \ForEach{$f \in$ Features of $C$}{
        \If{Difference in $f$ between $C$ and $h$ flips the outcome}{
            Add $f$ to $D$
        }
    }
    \textbf{return} $D$
}
\end{algorithm}

The counterfactual explanation (\Cref{lst:counterfactual_explanation}) illustrates what \textit{minimally} needs to change for an applicant to be eligible for a loan. 
In doing so, knowledge engineers can assess whether rules are too restrictive or lenient by comparing the outcomes of similar cases.

\begin{lstlisting}[
    float=htbp,
    caption={Counterfactual explanation suggesting actionable changes for \texttt{applicant1} to achieve loan eligibility by comparing their attributes with those of a successful applicant (\texttt{applicant3}).},
    label={lst:counterfactual_explanation},
    basicstyle=\scriptsize\ttfamily,
    breaklines=true,
    columns=fullflexible,
    frame=single,
    backgroundcolor=\color{gray!10},
    xleftmargin=1em,
    xrightmargin=0.5em,
    language={}
]
To change the outcome for applicant1 has Loan Eligibility: Not Eligible, you could look at these examples:

applicant3 has Loan Eligibility: Eligible because:
  - Their applicant3 has DTI Ratio: 0.2 while your applicant1 has DTI Ratio: 0.4
  - Their applicant3 has Monthly Debt: 1000.0 while your applicant1 has Monthly Debt: 2000.0
  - Their applicant3 has Credit Score: 700 while your applicant1 has Credit Score: 680
\end{lstlisting}

\begin{figure*}[t]
    \centering
    \begin{subfigure}[b]{0.22\textwidth}
        \includegraphics[width=\textwidth]{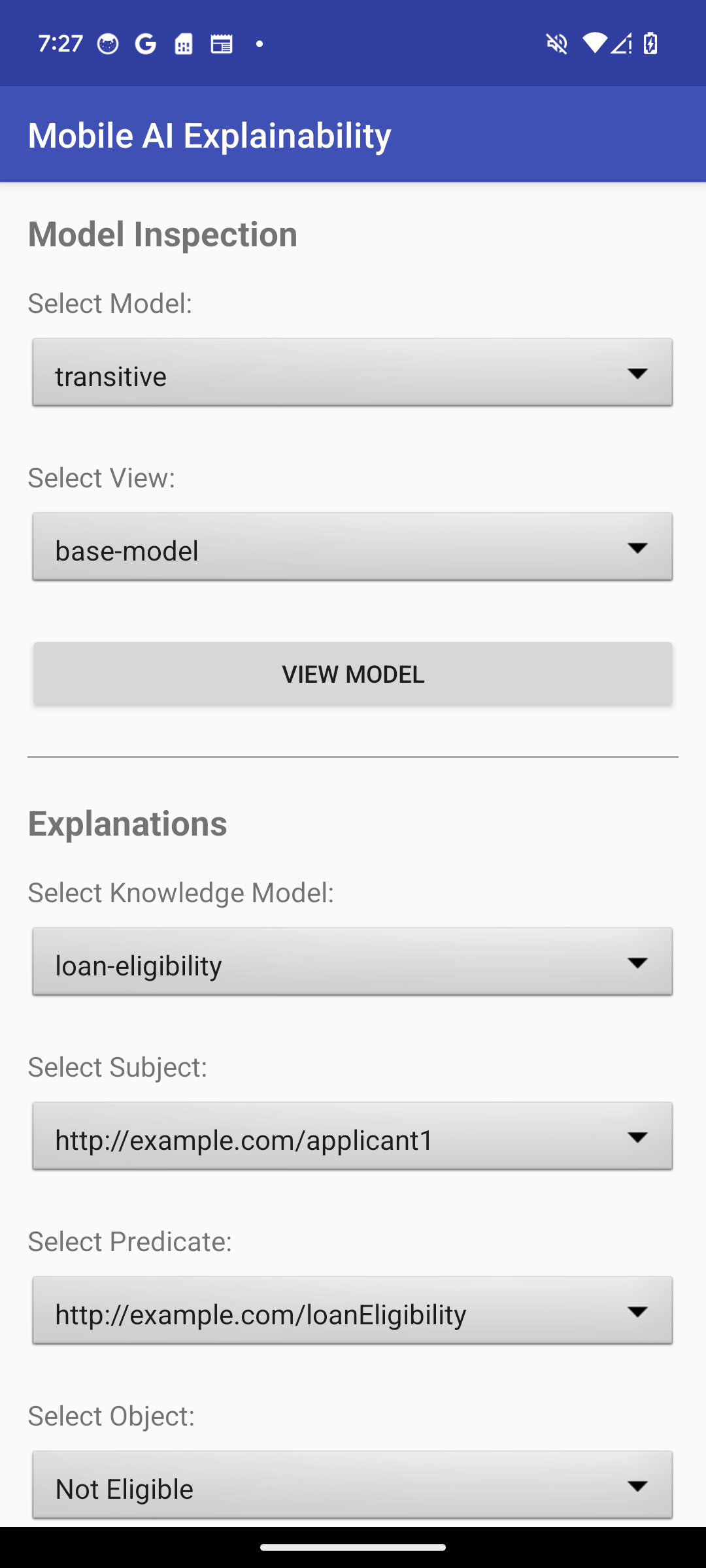}
        \caption{Model Inspection Pickers}
    \end{subfigure}
    \hfill
    \begin{subfigure}[b]{0.22\textwidth}
        \includegraphics[width=\textwidth]{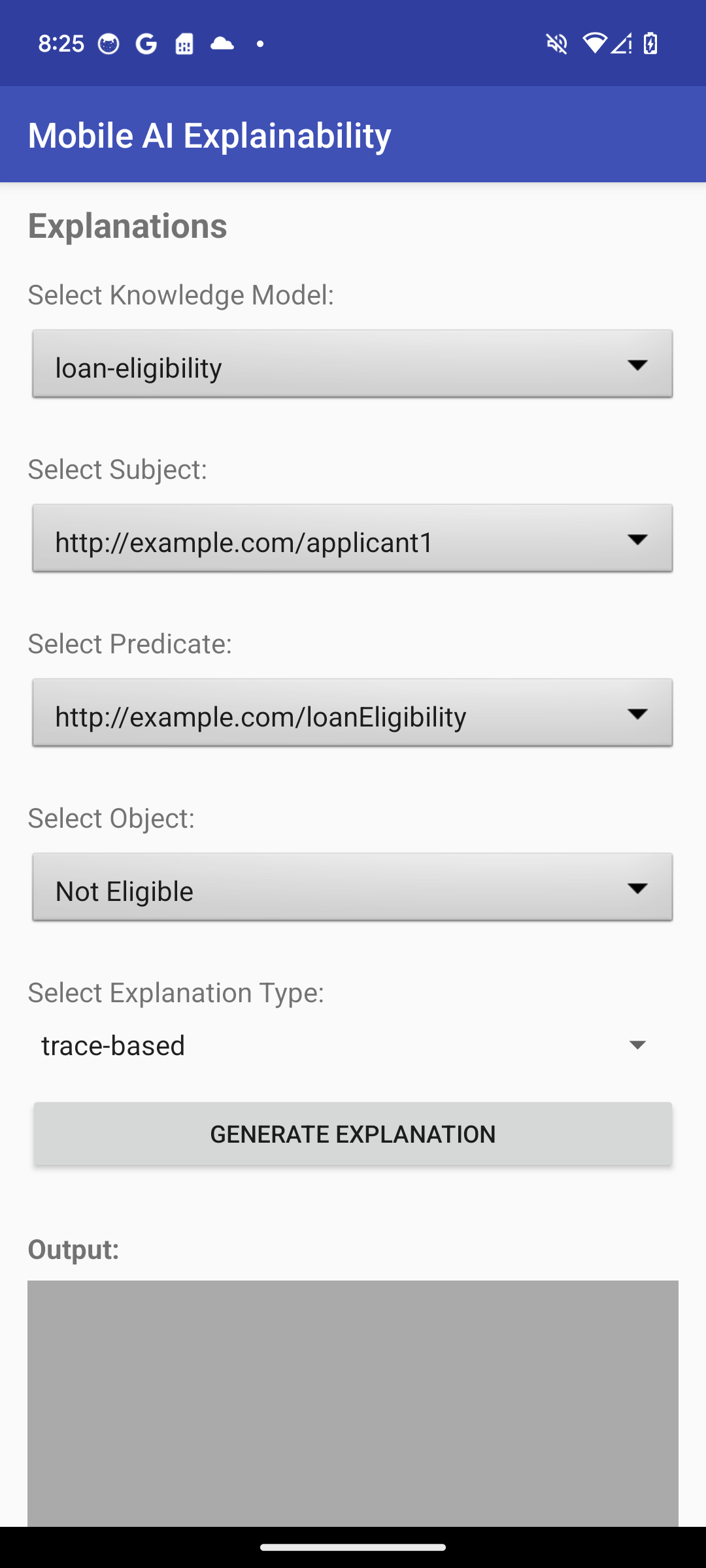}
        \caption{Explanation Input Pickers}
    \end{subfigure}
    \hfill
    \begin{subfigure}[b]{0.22\textwidth}
        \includegraphics[width=\textwidth]{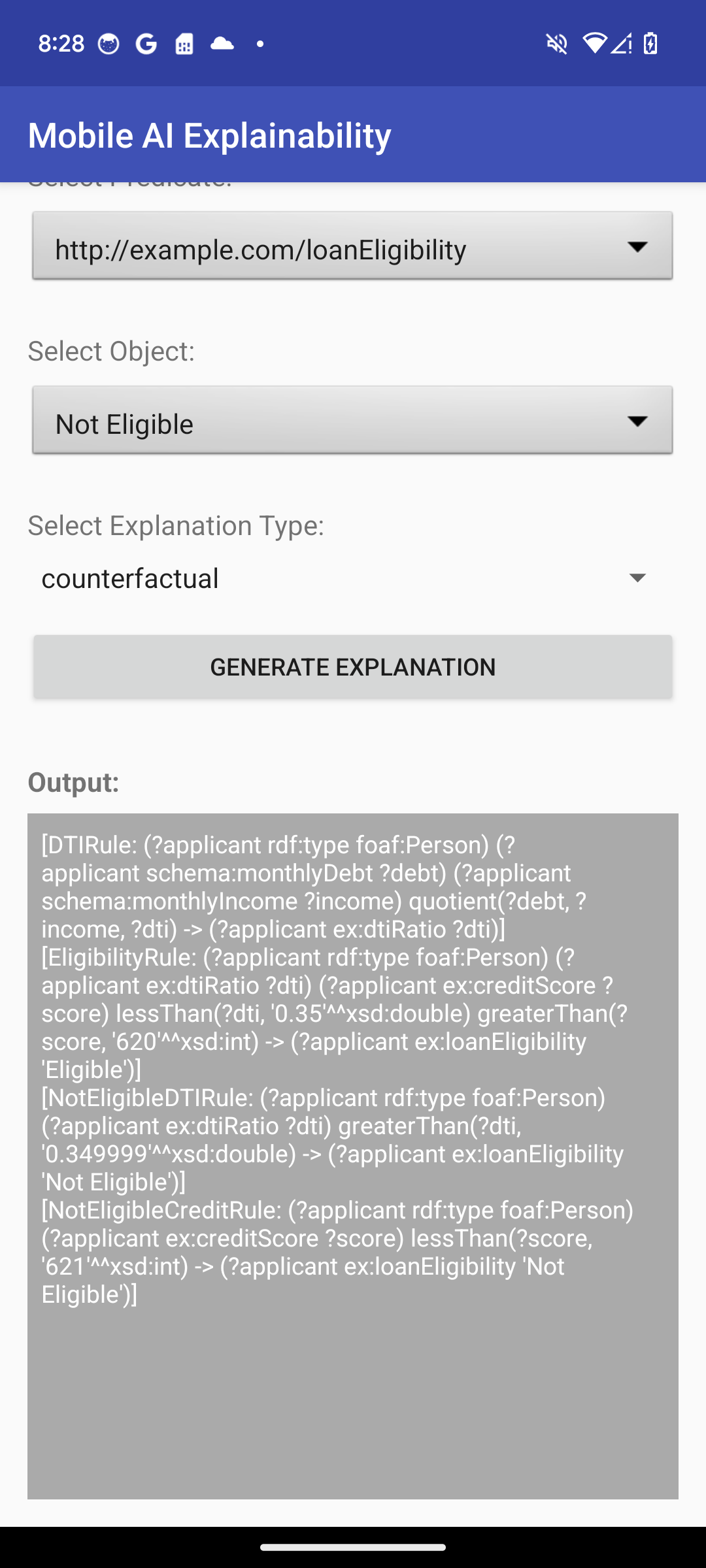}
        \caption{Loan Eligibility Model Rules}
    \end{subfigure}
    \hfill
    \begin{subfigure}[b]{0.22\textwidth}
        \includegraphics[width=\textwidth]{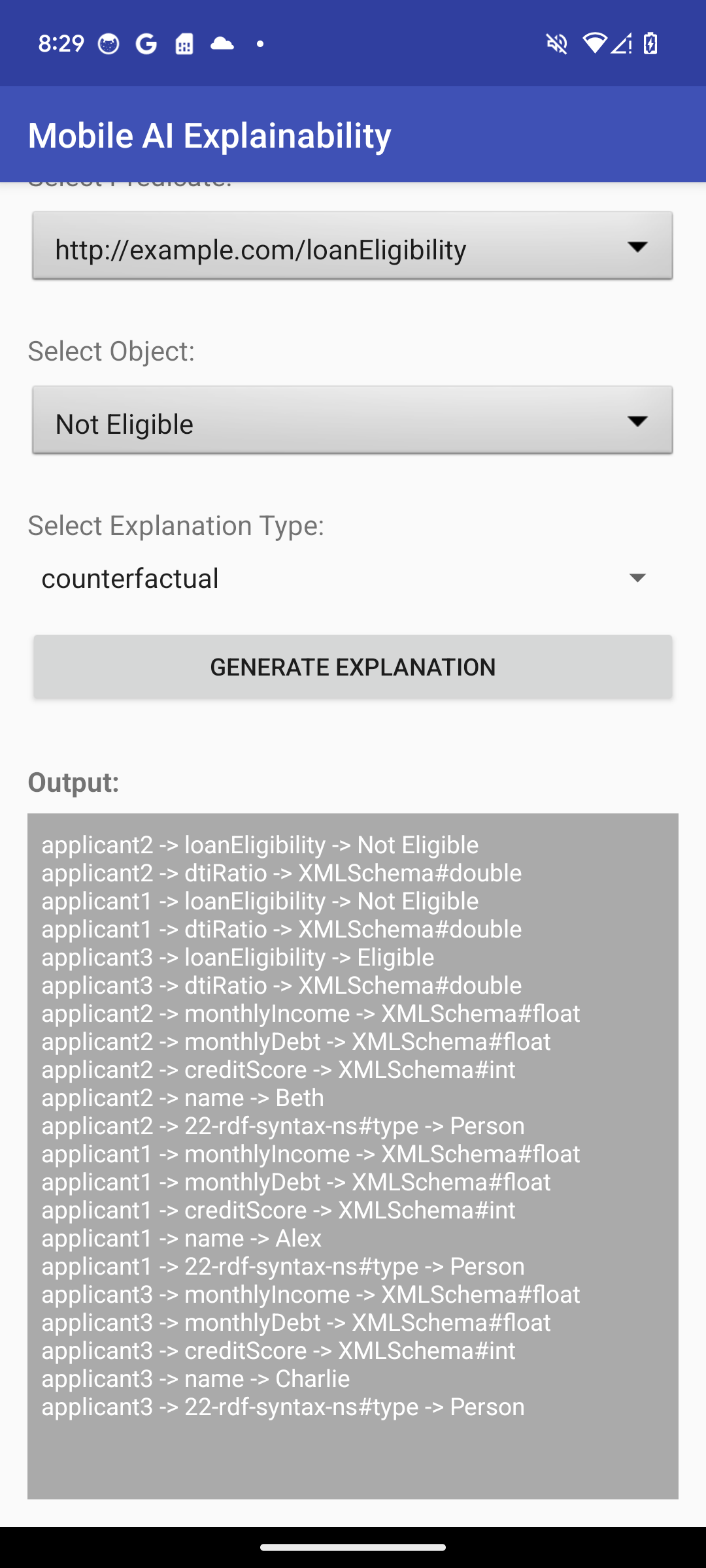}
        \caption{Inference Model}
    \end{subfigure}

    \vskip\baselineskip

    \begin{subfigure}[b]{0.22\textwidth}
        \includegraphics[width=\textwidth]{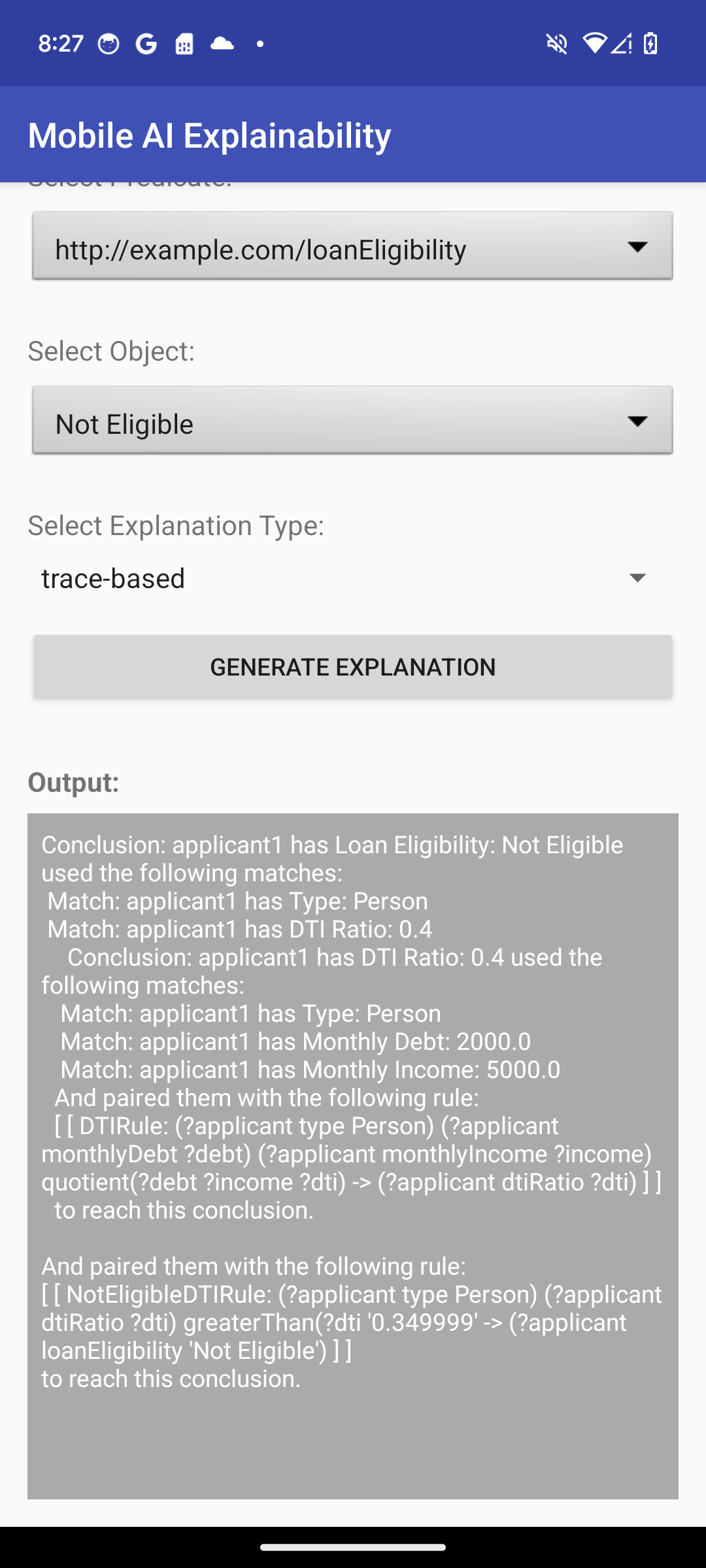}
        \caption{Loan Eligibility Trace-Based Explanation}
    \end{subfigure}
    \hfill
    \begin{subfigure}[b]{0.22\textwidth}
        \includegraphics[width=\textwidth]{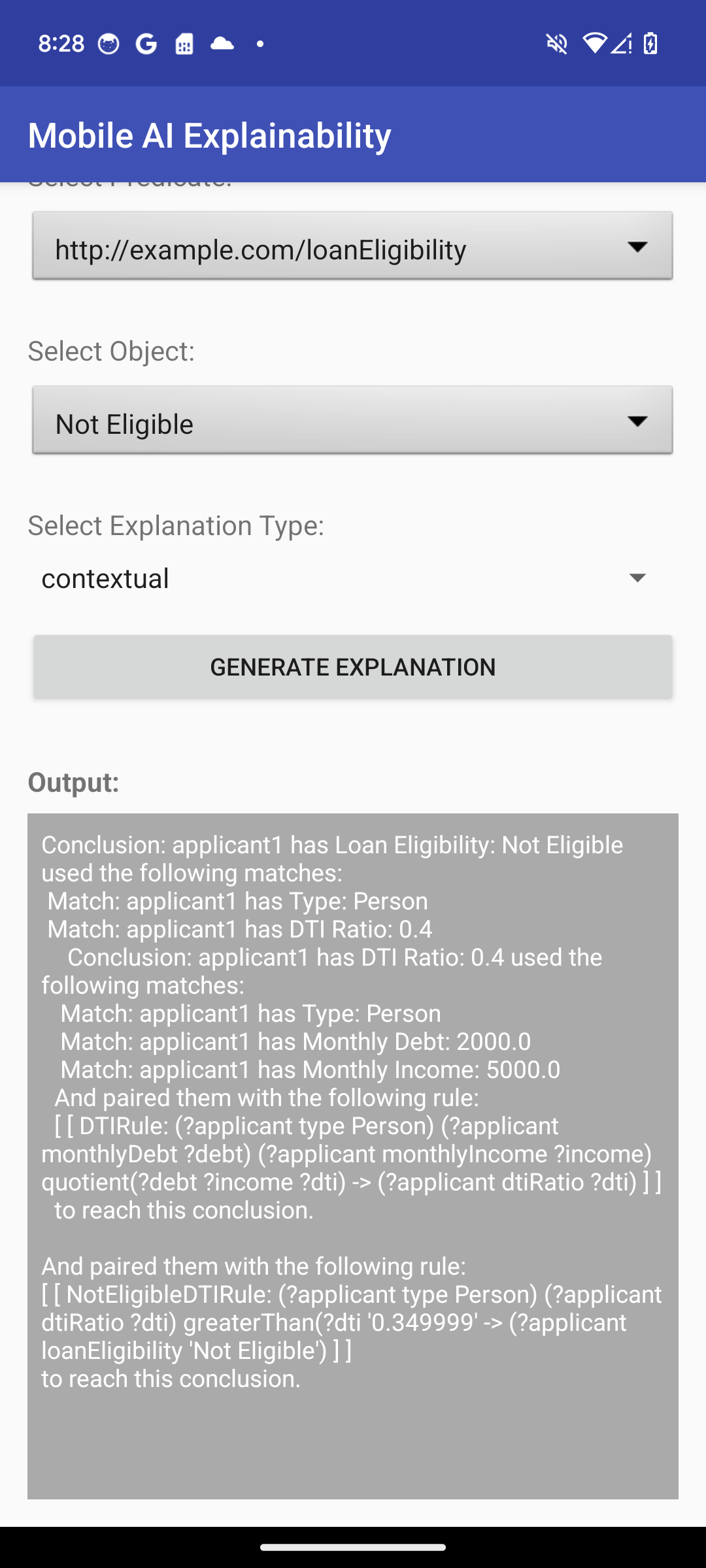}
        \caption{Loan Eligibility Contextual Explanation}
    \end{subfigure}
    \hfill
    \begin{subfigure}[b]{0.22\textwidth}
        \includegraphics[width=\textwidth]{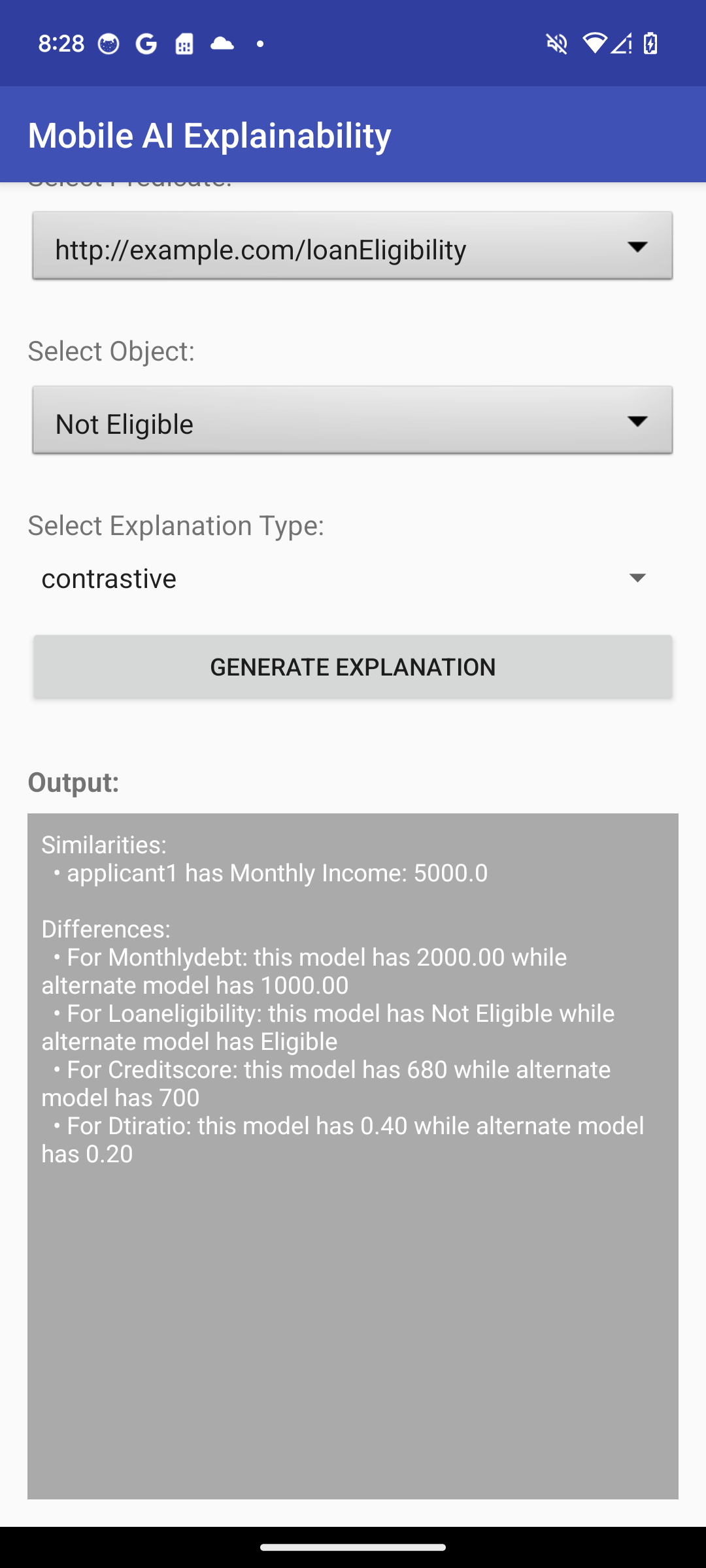}
        \caption{Loan Eligibility Contrastive Explanation}
    \end{subfigure}
    \hfill
    \begin{subfigure}[b]{0.22\textwidth}
        \includegraphics[width=\textwidth]{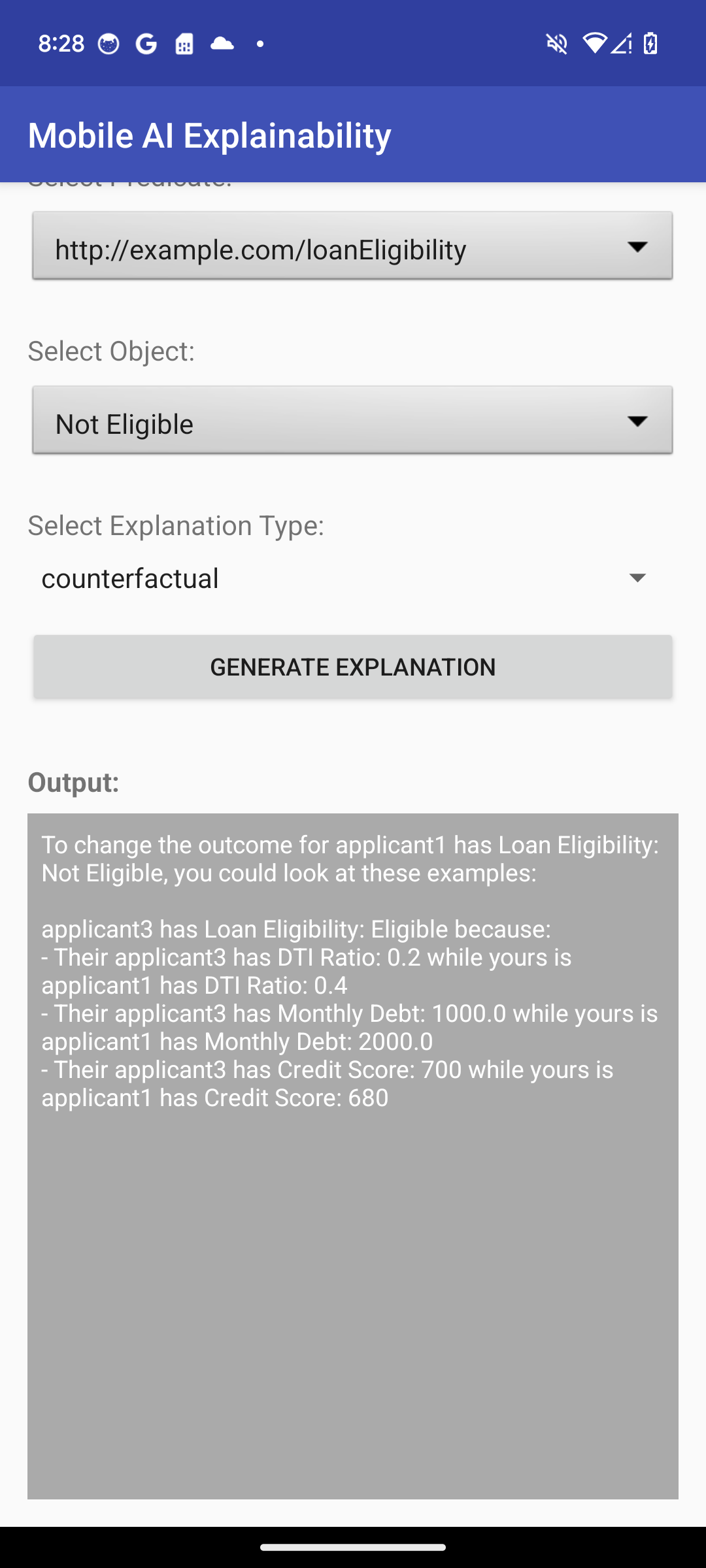}
        \caption{Loan Eligibility Counterfactual Explanation}
    \end{subfigure}

    \caption{Android App Screenshots}
    \label{fig:android_app_screenshots}
\end{figure*}

\section{Integration with MIT App Inventor for Rule Testing}
\label{sec:integration_testing}

The explanation component has been integrated into the MIT App Inventor Punya framework to facilitate the testing and validation of rules across various domain models (\Cref{fig:android_app_screenshots}). 

The test application consists of two primary sections that support the evaluation of rules and explanations: \textbf{Model Inspection} and \textbf{Explanations}. These sections enable users to interact with different domain models, inspect underlying rules and facts, and generate explanations for inferred conclusions.

The \textbf{Model Inspection} section allows users to explore the foundational elements of a reasoning system for three pre-loaded domain models, which currently include transitive reasoning, dietary recommendations, and loan eligibility (our running example). Users can select the domain model and the type of information they wish to inspect, including:
\begin{itemize}
    \item Base model facts
    \item Rules applied in the reasoning process
    \item Inference model facts
\end{itemize}

Once a model and information type are selected, users can view the output in the application’s interface along with the rules and facts. 

The \textbf{Explanations} section lets users test the explanation component’s functionality directly. Users can select:
\begin{itemize}
    \item A pre-loaded domain model
    \item A reasoned statement or triple (subject, predicate, object) to be explained
    \item The type of explanation desired (trace-based, contextual, contrastive, or counterfactual)
\end{itemize}

Upon selecting these parameters, the app generates an explanation by querying the inference model to describe how the selected inferred triple was created. The explanation is then displayed in the output box, providing knowledge engineers with insights into the reasoning process.

This integration highlights the practicality of the explanation component for real-world applications, aligning it with educational and research-oriented goals of MIT App Inventor Punya. 
In particular, it provides a user-friendly interface for testing and validating rules in different domain models by enabling:
\begin{itemize}
    \item Rule debugging and validation through detailed model inspection.
    \item Evaluation of different explanation types for various reasoning scenarios.
    \item Accessibility of rule-based reasoning and explanation capabilities on mobile platforms.
\end{itemize}


\section{Discussion}
\label{sec:discussion}

The use case in \Cref{sec:use_cases} highlights the practical utility of the explanation component in enhancing rule quality. 
\begin{enumerate*}[label=(\roman*)]
    \item \textbf{Trace-Based:} Enables a comprehensive understanding of reasoning chains, facilitating thorough validation and debugging of rule interactions.
    \item \textbf{Contextual:} Enhances human readability by isolating the immediate rules and facts that led to a specific conclusion, providing a focused view of the reasoning process.
    \item \textbf{Contrastive:} Illuminates critical differences between cases, enabling engineers to refine thresholds, address inconsistencies, and improve fairness.
    \item \textbf{Counterfactual:} Explores ``what-if" scenarios to test rule sensitivity and suggest actionable improvements, aligning outcomes with user goals.
\end{enumerate*}

By integrating these perspectives, the framework empowers knowledge engineers to systematically debug, validate, and refine rules. This iterative process strengthens the reliability and interpretability of the reasoning system, allowing its rules to serve as a robust foundation for the Web of Data. To the best of our knowledge, this paper is among the first to investigate the use of diverse explanation types as a tool for refining rules in knowledge-based reasoning systems. This novel approach shifts the focus from traditional computational correctness checks to explanation-driven, human-centric rule validation.

Regarding the distinction between trace-based and contextual explanations: 
while both contribute to understanding the reasoning process, contextual explanations can be viewed as a targeted application of trace-based methods. By focusing solely on the final stage of an inference chain, contextual explanations enhance human readability without overwhelming users with details. 

Beyond debugging, this explanation-driven framework has educational potential. By providing clear and structured insights into reasoning processes, the framework can be a tool for teaching rule crafting and knowledge graph development. Moreover, domain experts can utilize these tools to evaluate the accuracy and comprehensiveness of the rules they author. Aligning rule quality assessments with intrinsic quality dimensions (e.g., accuracy, consistency) and contextual ``fit for use" requirements promotes responsible AI governance and ensures alignment with real-world standards of data quality.

The framework has limitations that suggest avenues for future research. Currently, it supports four explanation types derived from the explanation ontology~\cite{chari2020explanation}. Expanding the framework to incorporate additional explanation types from the ontology could enhance its applicability across diverse scenarios. Another area for improvement is the natural language output of explanations. Simplifying and refining these outputs would make explanations more accessible, particularly for non-technical users, who are the target audience of MIT App Inventor.
Future work could also involve conducting user studies to evaluate the impact of different explanation types on usability, trust, and decision-making in real-world applications. Such studies would provide valuable feedback for refining the framework and its integration into scalable reasoning systems. 
These advancements could further 
establish explainability as a cornerstone for responsible, high-quality reasoning systems.

\section{Conclusion}
\label{sec:conclusion}

As the Web of Data continues to expand, we argue that the interconnection between data quality and explainability is a foundational aspect of building reliable and trustworthy knowledge-based systems. Rules are the cornerstone of reasoning systems, shaping the usability, fairness, and reliability of their derived insights. Addressing these challenges, this paper introduces an explanation framework 
that integrates diverse explanation types, including trace-based, contextual, contrastive, and counterfactual, offering a multifaceted toolkit for debugging, refining, and validating rules. By incorporating the \emph{Explainer Component} into the MIT App Inventor Punya framework, we aim to demonstrate the practical value of explanation-driven approaches in bridging the gap between complex reasoning systems and user-centric transparency. 

This framework elevates the quality of rules and lays a foundation for responsible, data-driven decision-making in scenarios where high-quality, robust reasoning is paramount. Future research could extend the framework by incorporating additional explanation types, improving natural language outputs for greater accessibility, and conducting empirical studies to assess the impact of explanations on system usability, trust, and decision-making. These advancements may validate explainability as a cornerstone of transparent, high-quality, and trustworthy knowledge-based reasoning systems, advancing the vision of a responsible and reliable Web of Data.

\section*{Resource Availability}
\label{sec:resource_availability}

GitHub repository with the code for the explanation framework and the rules used in the use case is available at \url{https://github.com/brains-group/androidReasoningPunya}. 




\balance
\bibliographystyle{ACM-Reference-Format}
\bibliography{references}






\end{document}